\documentclass[10pt,twocolumn,letterpaper]{article}
\usepackage[pagenumbers]{iccv}
\usepackage{multirow}
\newcommand{\qheading}[1]{\vspace{5pt}\mbox{\textbf{#1}\;}}

\definecolor{iccvblue}{rgb}{0.21,0.49,0.74}
\usepackage[pagebackref,breaklinks,colorlinks,allcolors=iccvblue]{hyperref}
\def\paperID{****}
\def\confName{ICCV}
\def\confYear{2025}
\newcommand{\model}[0]{Dense Policy\xspace}
\usepackage{colortbl} 
\usepackage{bm}
\usepackage{float} 

\title{Dense Policy: Bidirectional Autoregressive Learning of Actions}

\author{Yue Su$^{*1,2,\ddagger}$, Xinyu Zhan$^{*1}$, Hongjie Fang$^{1}$, Han Xue$^{1}$, \\ Hao-Shu Fang$^{1}$,Yong-Lu Li$^{1,3}$, Cewu Lu$^{1,3}$, and~ Lixin Yang$^{1,\dagger}$ \\
{\small {$^{1}$Shanghai Jiao Tong University}~
{$^{2}$Xidian University}~
{$^{3}$Shanghai Innovation Institute}}
}

\newcommand{\customfootnotetext}[2]{{
  \renewcommand{\thefootnote}{#1}
  \footnotetext[0]{#2}}}

\begin{document}

\twocolumn[{
    \maketitle
    \begin{center}
    \centering
    \captionsetup{type=figure}
    \includegraphics[width=\textwidth]{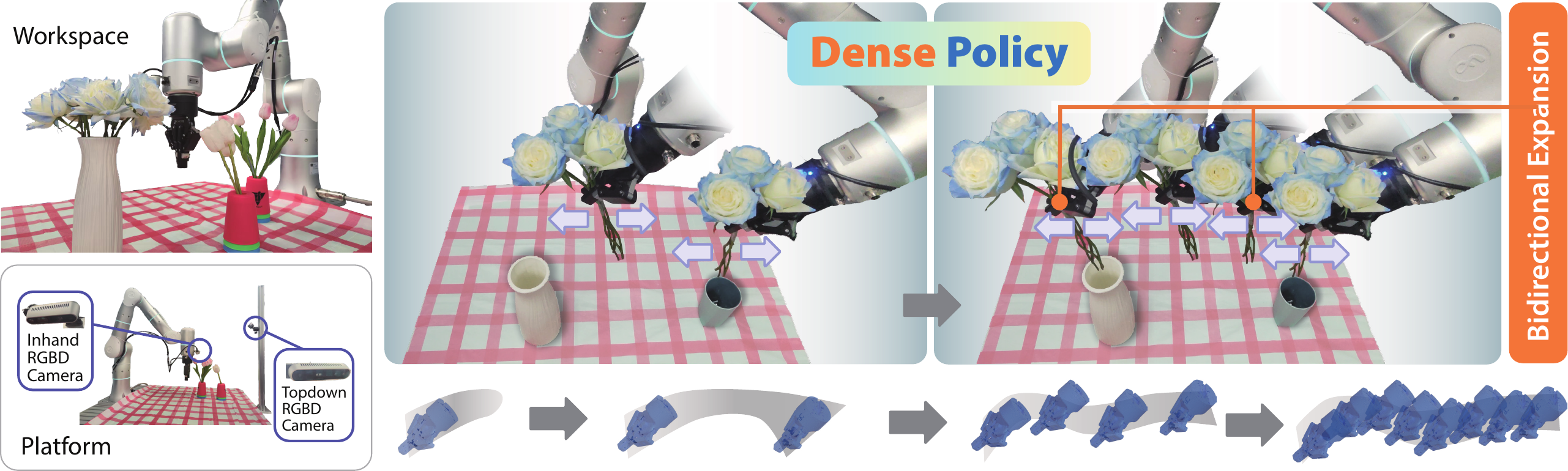}
    \captionof{figure}{\textbf{\model: a robot policy model that generates raw robot actions in an autoregressive manner}. During inference, \model performs bidirectional expansion on the current level of sparse action keyframes to obtain a denser action sequence.  }
    \vspace{5pt}
    \label{fig:front}
    \end{center}
}]
\customfootnotetext{$*$}{Equal contribution. $\dagger$ Corresponding author.}
\customfootnotetext{$\ddagger$}{This work is done while Y. Su is a research intern at Shanghai Jiao Tong University.}

\begin{abstract}
Mainstream visuomotor policies predominantly rely on generative models for holistic action prediction, while current autoregressive policies, predicting the next token or chunk, have shown suboptimal results. This motivates a search for more effective learning methods to unleash the potential of autoregressive policies for robotic manipulation. This paper introduces a bidirectionally expanded learning approach, termed Dense Policy, to establish a new paradigm for autoregressive policies in action prediction. It employs a lightweight encoder-only architecture to iteratively unfold the action sequence from an initial single frame into the target sequence in a coarse-to-fine manner with logarithmic-time inference. Extensive experiments validate that our dense policy has superior autoregressive learning capabilities and can surpass existing holistic generative policies. Our policy, example data, and training code will be publicly available upon publication. Project page: \url{https://selen-suyue.github.io/DspNet/}.
\end{abstract}    
\section{Introduction}
\label{sec:intro}

\begin{figure*}[!t]
  \centering
  \begin{minipage}{0.64\textwidth} 
    \includegraphics[width=\textwidth]{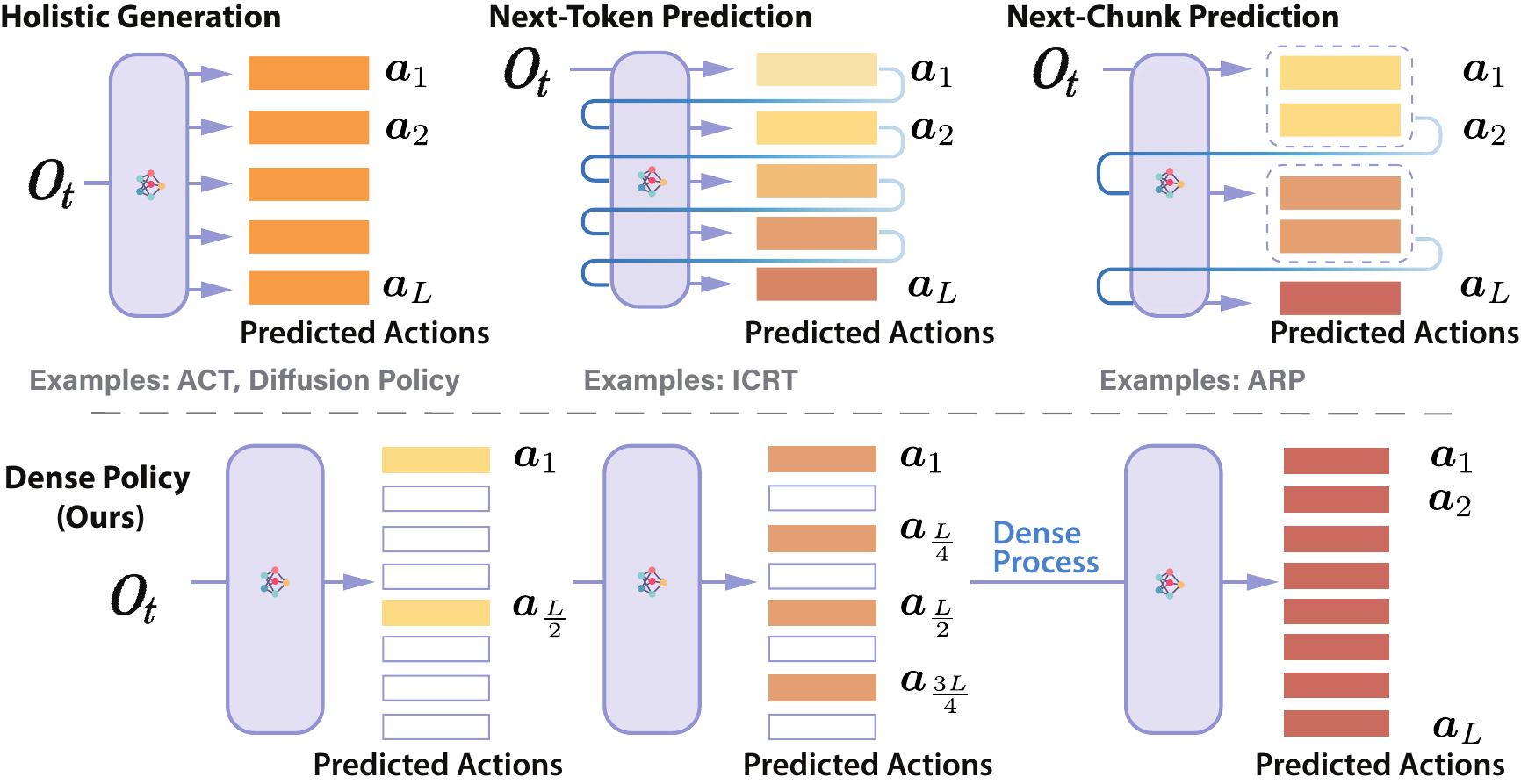}
    \caption*{\textit{\textbf{a)} Different Generation Paradigms}}
  \end{minipage}\hfill 
  \begin{minipage}{0.34\textwidth} 
    \includegraphics[width=\textwidth]{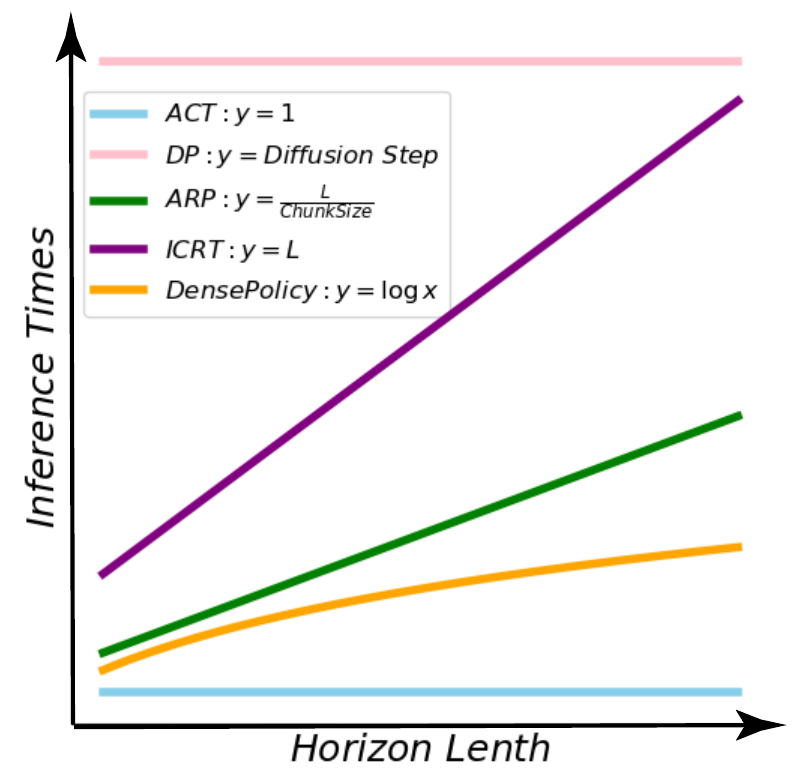} 
    \caption*{\textit{\textbf{b)} Inference Time Complexity}}
  \end{minipage}
    \caption{\textbf{The Difference between \model and Existing Policies.}
    Holistic generation policies model the joint distribution of sequences to produce all actions over a horizon. 
    For ACT, this requires a single variational inference step, whereas DP requires multiple diffusion steps. 
    In autoregressive policies, Next-Token prediction recursively generates actions for subsequent time steps, while Next-Chunk generates actions in segments, both achieving this within linear complexity. 
    In contrast, \model employs a bidirectional extension approach to achieve hierarchical action prediction, completing generation within logarithmic complexity.
    }
\label{relat}
\end{figure*}

Imitation Learning (IL) has become a prominent focus in robotic manipulation for its capability in robot learning. Its effectiveness is demonstrated both in downstream policies~\cite{ACT,DP,DP3,RISE,cage} tailored to specific tasks, and in Vision-Language-Action (VLA) models~\cite{openvla,octo,pi0} designed for general-purpose scenarios. Among those IL methods, holistic generative policies ~\cite{ACT,DP,DP3,RISE} prevail as the dominant approach in manipulation for their precise predictions within continuous action spaces.

Recent research has sought to enhance imitation learning by introducing autoregressive paradigm~\cite{attention,GPT,BERT} for its powerful sequence modeling capabilities and impressive performance under scaling~\cite{scalinglaw}.
While effective in language and vision, this paradigm faces significant challenges in action prediction.
Actions represent a more complex search space~\cite{continuous} with sparser data~\cite{rt1,rt2,oxe}, making it difficult to achieve precise predictions.
The traditional next-token prediction approach~\cite{attention}, commonly used in autoregressive models, is limited in capturing long-term dependencies within action sequences, leading to suboptimal performance~\cite{ManipLLM,vima} than holistic generative policies~\cite{ACT,DP}.
Recent efforts~\cite{CARP} address this limitation by introducing multi-scale prediction inspired by Visual Autoregressive Modeling~\cite{VAR}, while discrete actions and codebook construction make precise prediction and training harder~\cite{mar}.

The above works prompt a question: \textit{has the full potential of autoregressive models been realized for action generation?} 
Consider human manipulation: rather than sequentially inferring the action trajectory step-by-step, humans often envision a few keyframes that span the entire task execution and subsequently refine the operational process~\cite{computational,hierarchical,architecture}.
This resembles the concept of the receptive field in vision~\cite{cnn}, suggesting a coarse-to-fine perceptual process governing human action execution.

Building on this intuition, we propose \textbf{\model}, an autoregressive policy model features bidirectional expansion learning. As shown in \cref{fig:front},
\model generates precise predictions in raw continuous action spaces without a specially designed action tokenization mechanism.
Specifically, the model employs an encoder-only architecture to achieve hierarchical action prediction from observations.
The process begins with a constant vector representing the initial single-frame action.
Sparse keyframe actions are then derived via cross-attention with observation features embedded by the encoder.
Subsequently, a recursive ``Dense Process'' is applied: the sparse action sequence from the previous level is expanded bidirectionally through upsampling.
The upsampled actions are then refined via cross-attention with observation features in the same encoder, yielding a more fine-grained and precise action representation at the next level.
This process doubles the sequence length at each iteration. Thus, after a logarithmic number of recursions, the length of the generated action sequence reaches the predefined horizon, serving as the final representation for producing the action outputs.

We have implemented and tested both the 2D and 3D versions of \textbf{\model}, conducting extensive experiments across 11 tasks in 3 simulation benchmarks and 4 real-world tasks. These experiments compared 4 policies~\cite{DP,DP3,RISE,ACT} in 2D and 3D settings, yielding significant results.
We make the following key contributions:
\begin{itemize}
    \item We introduce a novel autoregressive policy that demonstrates superior performance over existing generative policies in both simulated and real-world tasks, encompassing both 2D and 3D scenarios.
    \item By introducing the Dense Process, we present the first bidirectional autoregressive learning approach for actions and demonstrate its superior task execution capabilities.
    \item We adopt an encoder-only architecture and demonstrate its advantages in terms of lightweight, rapid inference, and ease of training.
\end{itemize}

\section{Related Work}
\label{sec:related}

\subsection{Imitation Learning for Manipulation}
Imitation learning, especially behavior cloning~\cite{bc,ibc,bcz} has emerged as a powerful paradigm for robots to acquire complex skills from expert demonstrations.
Recent advances have witnessed a surge in the integration of generative models into the traditional behavior cloning framework.
Among the most prominent approaches are methods that employ Conditional Variational Autoencoders~\cite{vae}, such as ACT~\cite{ACT}, and diffusion model~\cite{DDIM,DDPM}-based policies~\cite{DP,DP3,RISE} for both 2D and 3D action spaces.
These methods typically condition on observations and model the joint distribution of the target action sequence in a generative manner, enabling the holistic generation of target action trajectories.  

Recent research has increasingly focused on token-wise incremental generation for robot policy learning. 
For instance, ICRT~\cite{icrt} applies the next-token prediction paradigm to action generation, achieving step-by-step action generation rather than holistic generation.
ARP~\cite{arp} combines target actions into chunks for multi-token next-step prediction, addressing the limitations of single-token prediction's attention span.
CARP~\cite{CARP}, inspired by VAR~\cite{VAR}, uses a multi-scale VQ-VAE~\cite{vqvae} approach to predict complete action sequences at multiple scales and progressively approaches the ground truth in a residual manner.  

However, bidirectional dependencies between actions at different timesteps hinder effective capture via next-token prediction.
Moreover, pixel-like discrete multi-scale reconstruction hinders the accuracy requirements of the action.
The aforementioned issues prevent autoregressive policies from widespread use in real-world tasks.

\subsection{Multidirectional Autoregressive Model}
To enhance the global understanding of the target, many autoregressive models pursue a non-unidirectional learning paradigm.
BERT~\cite{BERT} pioneered bidirectional context learning through masked language modeling, which subsequently spurred significant advancements in the field of natural language processing~\cite{RoBERTa,SpanBERT,Ernie}.

More recently, autoregressive models in the image domain also have begun to depart from the conventional raster order generation, initiating a shift towards non-raster order approaches.
SAIM~\cite{saim} transcends the limitations of traditional raster-scan order by predicting image patches in a random sequence.
MAR~\cite{mar} demonstrates that fixed orders and discrete representations are unnecessary, and employs a diffusion loss to train a continuous representation model in a Masked Image Modeling style.
SAR~\cite{sar} constructs a unified framework that enables causal learning to adapt to arbitrary sequence orders and output intervals, thereby providing a more flexible and powerful approach to autoregressive image modeling.

The aforementioned works motivate us to explore bidirectional generation methods for action sequences to obtain more precise action representations.
Diverging from these approaches, we replace the conventional masked prediction with a bidirectional expanding strategy.

\begin{figure*}[!t]
  \centering
    \includegraphics[width=\textwidth]{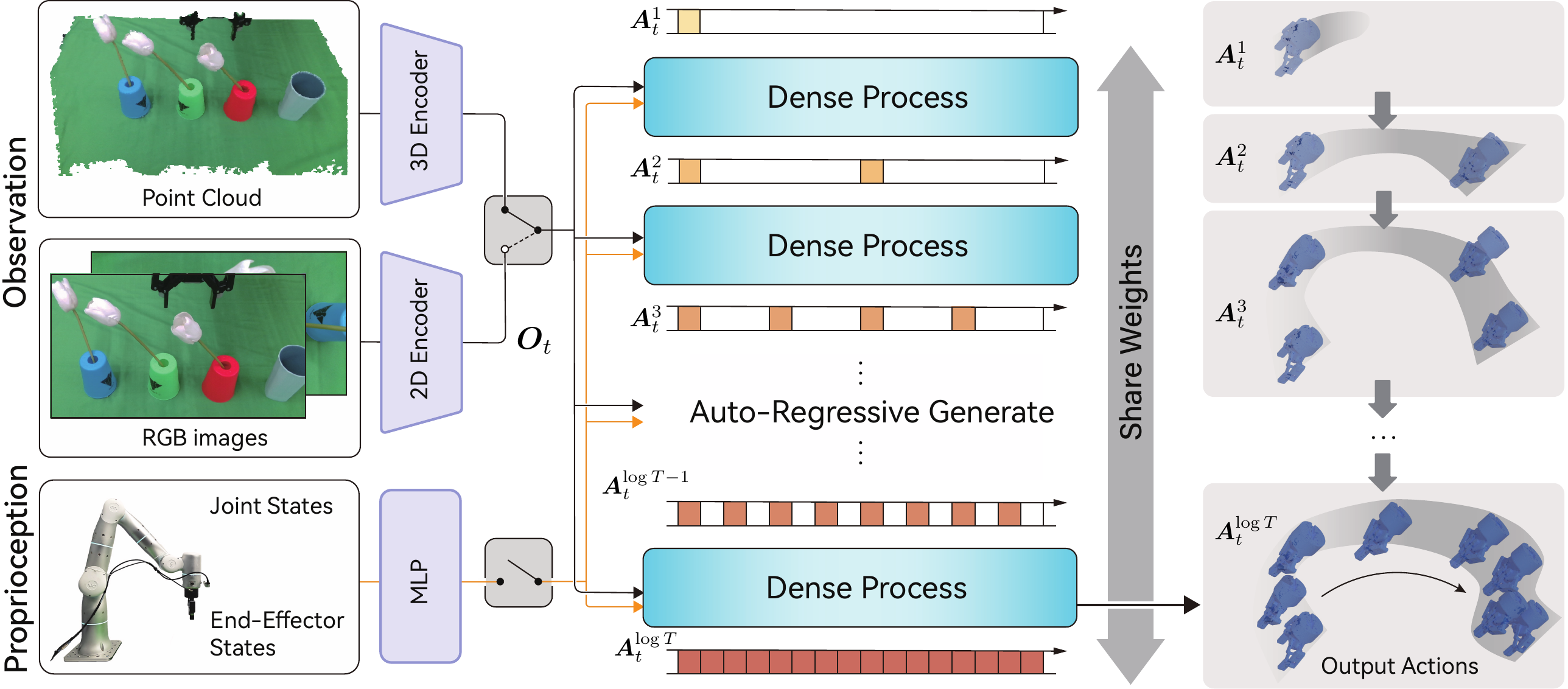}
    \caption{\textbf{Overview of \model}. \model accepts visual inputs in different modalities and optional robot proprioception. It employs a unified encoder to perform cross-attention between hierarchical action representations and observation features. This facilitates a bidirectionally expanding dense process. During each dense process level, the actions, initially represented as sparse keyframes, are progressively infilled and refined into a complete predicted sequence, leading to a coarse-to-fine generation procedure.}
    \label{pipeline}
\end{figure*}
\section{Method}

\subsection{Problem Formulation}
We aim to facilitate bidirectional learning within autoregressive policies, thereby achieving a multi-scale ``receptive field'' approximation for the action modality.
This involves initially generating a sparse set of keyframe actions spanning the trajectory and subsequently performing a gradual infilling and refining process to construct the complete dense action sequence.
This coarse-to-fine generation paradigm is thus termed \model. The pipeline of our method is shown in \cref{pipeline}. Additionally, in~\cref{relat}, we elucidate its differences from mainstream policies.

Formally, let $\bm{O}_t$ denote the observation at time $t$, which may consist of visual inputs (\eg, RGB images or point clouds) and optional proprioceptive information.
The objective is to predict a future action sequence $\bm{A}_{t:t+T} = \{\bm{a}_{t}, \bm{a}_{t+1}, ..., \bm{a}_{t+T-1}\}$ over a horizon of $T$ time steps.
In this work, the action $\bm{a}_t$ represents the TCP pose of the robot's end-effector.
Under the standard generative behavior cloning framework~\cite{ACT,DP}, this is typically achieved by maximizing the likelihood of the actions given the observation by $\bm{P}(\bm{A}|\bm{O})$.
While \model decomposes the overall objective $\bm{A}$ into a hierarchy of representations: $\bm{A}^1,\bm{A}^2... \bm{A}^{\log_{2} T}$, where  
\begin{equation}
    \bm{A}^n=\{\bm{a}_{t+i}^n|\ i\ mod \ \frac{T}{2^n}= 0,\ \ i \in \mathbb{N}_{<T}\}
    \label{denselevel}
\end{equation}
represents an intermediate action sequence at a different level of granularity at the $n$-th level, emphasizing that the layer has generated a coarse representation of the actions within those $n$ time steps, which will be progressively refined and optimized in subsequent processes. 
It should be noted that $\bm{A}^{0}$ is $\bm{\emptyset}$ according to~\cref{denselevel} which represents the initial constant action vector. Specifically, we set \(\bm{A}^{0} = \mathbf{0}\) to provide an unbiased starting point for the iterative refinement process.
It allows the model to learn the action sequence purely based on the observation, without any prior assumptions about the initial action.
Thus, our model can be formulated as:
\begin{equation}
    \bm{P}(\bm{A}|\bm{O}) = \prod_{i=1}^{n}\bm{P}(\bm{A}^i|\bm{A}^{i-1},\bm{A}^{i-2},\dots,\bm{A}^0,\bm{O}).
\end{equation}

\subsection{Observation Encoder} 
While the primary contribution of \model lies in the design of the action head, rather than a specific visual backbone, it is easily integrable with various visual backbones and adaptable to different visual inputs.
Inspired by \cite{DP,RISE}, \model uses ResNet18~\cite{resnet} with GroupNorm~\cite{groupnorm} as the default 2D visual encoder and the sparse convolutional network~\cite{spatioenc} as the default 3D encoder unless otherwise specified. 
We demonstrate in subsequent experiments that \model can seamlessly inherit other visual encoding methods~\cite{DP3} without compromising the performance of the Dense Policy action head. 

In scenarios requiring proprioceptive awareness, \model randomly masks portions of the robot's end-effector pose during training.
This mitigates the risk of the model developing a bias towards memorizing actions associated with fixed positions, thereby enhancing generalization performance.
Proprioceptive information is consistently encoded into the feature space using MLP.

\subsection{Autoregressive Dense Policy}
After feature encoding, actions are recursively expanded in a hierarchical manner.
Specifically, each transition between hierarchical levels undergoes a dense process, illustrated in the center column in \cref{pipeline}.
Actions from the preceding level $\bm{A}^{n}$ are first expanded via linear upsampling.
The upsampled actions $\bm{A}_{up}^{n}$ at the current level are denoted as:
\begin{equation}
    \bm{A}_{up}^{n}=\{\tilde{\bm{a}}_{t+j}^n|\ j\ \mathrm{mod} \ \frac{T}{2^{n+1}}= 0,\ \ j \in \mathbb{N}_{<T}\}.
\end{equation}
Each element $\tilde{\bm{a}}_{t+j}^n$ in the upsampled actions is obtained from the following process:
\begin{equation}
    \tilde{\bm{a}}_{t+j}^n = \left\{\begin{array}{lr}
        \bm{a}_{t+j} & \text{if } j \ \mathrm{mod}\ \frac{T}{2^n} = 0 \\[10pt]
        \multicolumn{2}{l}{
            \frac{1}{2} \big(
                \bm{a}_{t+j-\frac{T}{2^{n+1}}} +
                \bm{a}_{t+j+\frac{T}{2^{n+1}}} 
            \big)
        } \\[5pt]
        \multicolumn{2}{r}{
            \quad \quad \quad
            \text{if } j \ \mathrm{mod}\ \frac{T}{2^n} \neq 0
        }  \\[10pt]
        \bm{a}_{t+T-\frac{T}{2^{n}}} & \text{if }j = T - \frac{T}{2^{n+1}}
    \end{array}\right.
    .
\end{equation}

Subsequently, \model embeds the observation features preprocessed by the visual backbone into a BERT Encoder~\cite{BERT}.
This embedding is then used to perform cross-attention with the upsampled actions from the previous level with 4 Encoder Layers, ultimately outputting the action representation for the next level $A^{n+1}$:
\begin{equation}
    \bm{A}^{n+1}= \mathbf{Enc}(\bm{A}^n_{up},\bm{O}).
\end{equation}
During the transformation process, the key actions from the higher level serve as prior knowledge to guide the prediction of more fine-grained actions at the subsequent level, illustrating the concept of a ``dense process''.
This process is repeated until the sequence is expanded to the target horizon length. The final action representation is projected onto the predicted action space via a linear layer. $L_2$ loss is used to supervise action predictions against ground-truth actions.

\section{Simulation Experiments}
\begin{table*}[!t]
    \centering
    \renewcommand\arraystretch{1.6}
    \setlength\tabcolsep{3pt}
    \setlength{\aboverulesep}{0pt}
    \setlength{\belowrulesep}{0pt}
    \footnotesize
    \begin{tabular}{c|cccccccccccc}
        \hline
        \multirow{2}{*}{\textbf{Method}}  & \multicolumn{2}{c}{\textit{\textbf{Adroit}}} & \multicolumn{2}{c}{\textit{\textbf{DexArt}}} & \multicolumn{7}{c}{\textit{\textbf{MetaWorld}}}&\multirow{2}{*}{\textbf{Avg}} \\
        \cmidrule(r){2-3}\cmidrule{4-5}\cmidrule(l){6-12}
        & \textit{Door} & \textit{Pen} & \textit{Laptop} & \textit{Toilet} & \textit{Bin Picking} & \textit{Box Close} & \textit{Hammer} & \textit{Peg Insert Side} & \textit{Disassemble} & \textit{Shelfplace} & \textit{Reach} \\
        \hline
        \cellcolor[HTML]{ecf8f8} 3D \model & \cellcolor[HTML]{ecf8f8}$\mathbf{72}\pm3$ & \cellcolor[HTML]{ecf8f8}$\mathbf{61}\pm0$ & \cellcolor[HTML]{ecf8f8}$\mathbf{85}\pm4$ & \cellcolor[HTML]{ecf8f8}$\mathbf{74}\pm3$ & \cellcolor[HTML]{ecf8f8}$\mathbf{47}\pm10$ & \cellcolor[HTML]{ecf8f8}$\mathbf{69}\pm8$ & \cellcolor[HTML]{ecf8f8}$\mathbf{100}\pm0$ & \cellcolor[HTML]{ecf8f8}$\mathbf{82}\pm4$ & \cellcolor[HTML]{ecf8f8}$\mathbf{98}\pm1$ & \cellcolor[HTML]{ecf8f8}$\mathbf{77}\pm4$ & \cellcolor[HTML]{ecf8f8}$\mathbf{31}\pm3$ & \cellcolor[HTML]{ecf8f8}$\mathbf{72}\pm4$\\
        DP3~\cite{DP3} & $62\pm4$ & $43\pm6$ & $81\pm2$ & $71\pm3$ & $34\pm30$ & $42\pm3$ & $76\pm4$ & $69\pm7$ & $69\pm4$ & $17\pm10$ & $24\pm1$ & $53\pm7$\\
        \hline
        \cellcolor[HTML]{ecf8f8} 2D \model & \cellcolor[HTML]{ecf8f8}$\mathbf{59}\pm8$ & \cellcolor[HTML]{ecf8f8}$\mathbf{65}\pm1$ & \cellcolor[HTML]{ecf8f8}$28\pm7$ & \cellcolor[HTML]{ecf8f8}$\mathbf{36}\pm8$ & \cellcolor[HTML]{ecf8f8}$\mathbf{25}\pm2$ & \cellcolor[HTML]{ecf8f8}$\mathbf{51}\pm3$ & \cellcolor[HTML]{ecf8f8}$\mathbf{86}\pm4$ & \cellcolor[HTML]{ecf8f8}$\mathbf{60}\pm7$ & \cellcolor[HTML]{ecf8f8}$\mathbf{71}\pm6$ & \cellcolor[HTML]{ecf8f8}$\mathbf{59}\pm6$ & \cellcolor[HTML]{ecf8f8}$\mathbf{27}\pm4$ & \cellcolor[HTML]{ecf8f8}$\mathbf{52}\pm5$\\
        DP~\cite{DP} & $37\pm2$ & $13\pm2$ & $\mathbf{31}\pm4$ & $26\pm8$ & $15\pm4$ & $30\pm5$ & $15\pm6$ & $34\pm7$ & $43\pm7$ & $11\pm3$ & $18\pm2$ & $25\pm5$\\
        \hline
    \end{tabular}
    \caption{\textbf{Detailed performance of \model\ and baselines in simulation tasks.} We report the mean of the top 5 success rates from three different random seeds.}
    \label{simtask}
    \vspace{-5pt} %
\end{table*}
\begin{figure*}[!t]
\centering
\begin{minipage}{0.245\textwidth}
  \centering 
  \includegraphics[width=\textwidth]{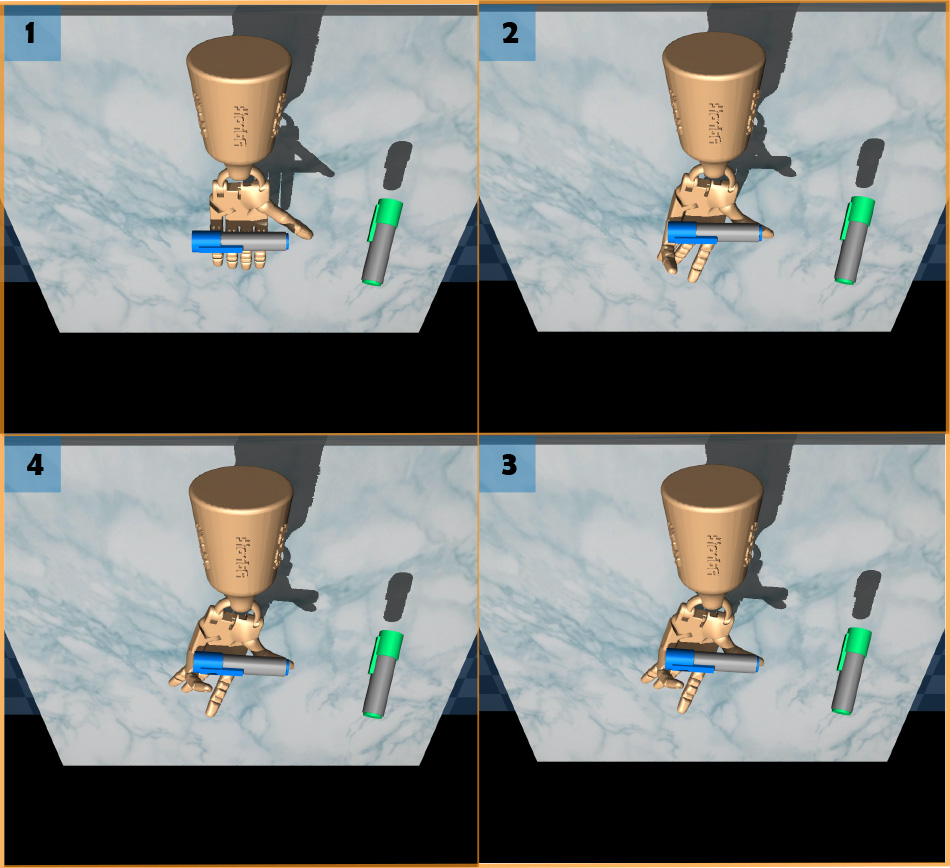}
  \caption*{\textbf{\textit{Adroit-Pen}}}
  \label{fig:pen}
\end{minipage}\hfill
\begin{minipage}{0.245\textwidth}
  \centering 
  \includegraphics[width=\textwidth]{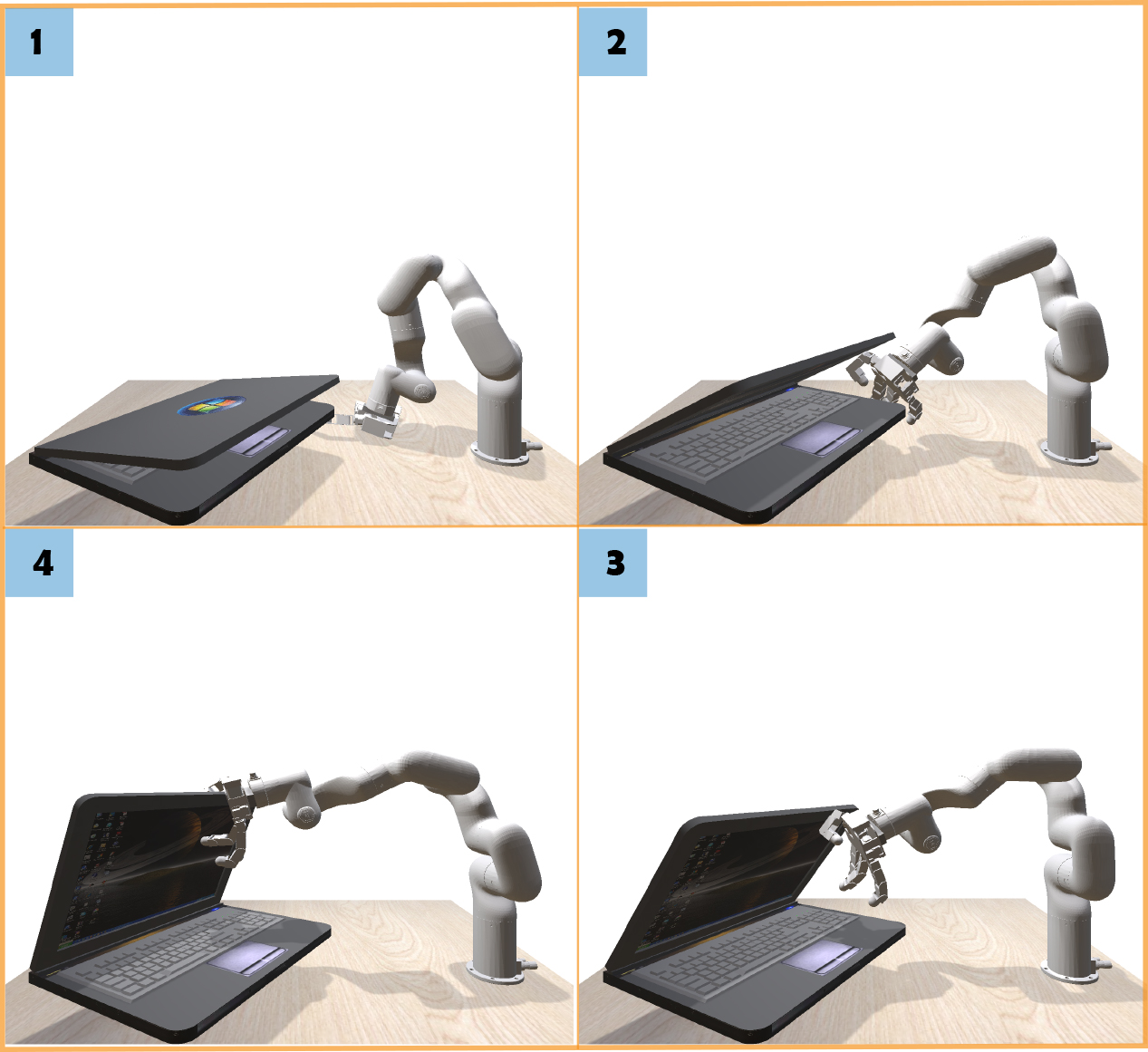}
  \caption*{\textbf{\textit{Dexart-Laptop}}}
  \label{fig:laptop}
\end{minipage}\hfill
\begin{minipage}{0.245\textwidth}
  \centering 
  \includegraphics[width=\textwidth]{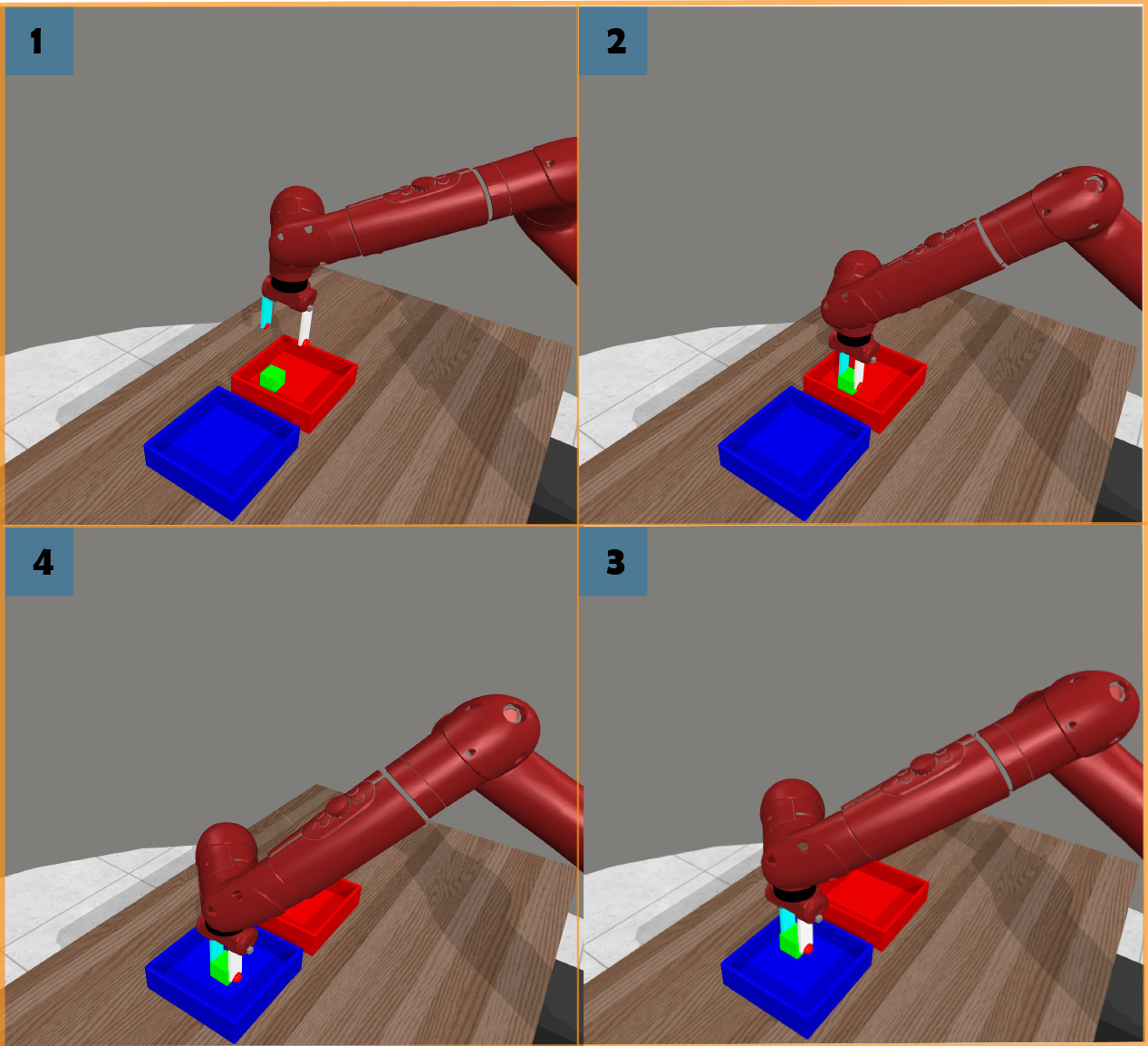}
  \caption*{\textbf{\textit{Metaworld-Bin Picking}}}
  \label{fig:binpicking}
\end{minipage}\hfill
\begin{minipage}{0.245\textwidth}
  \centering 
  \includegraphics[width=\textwidth]{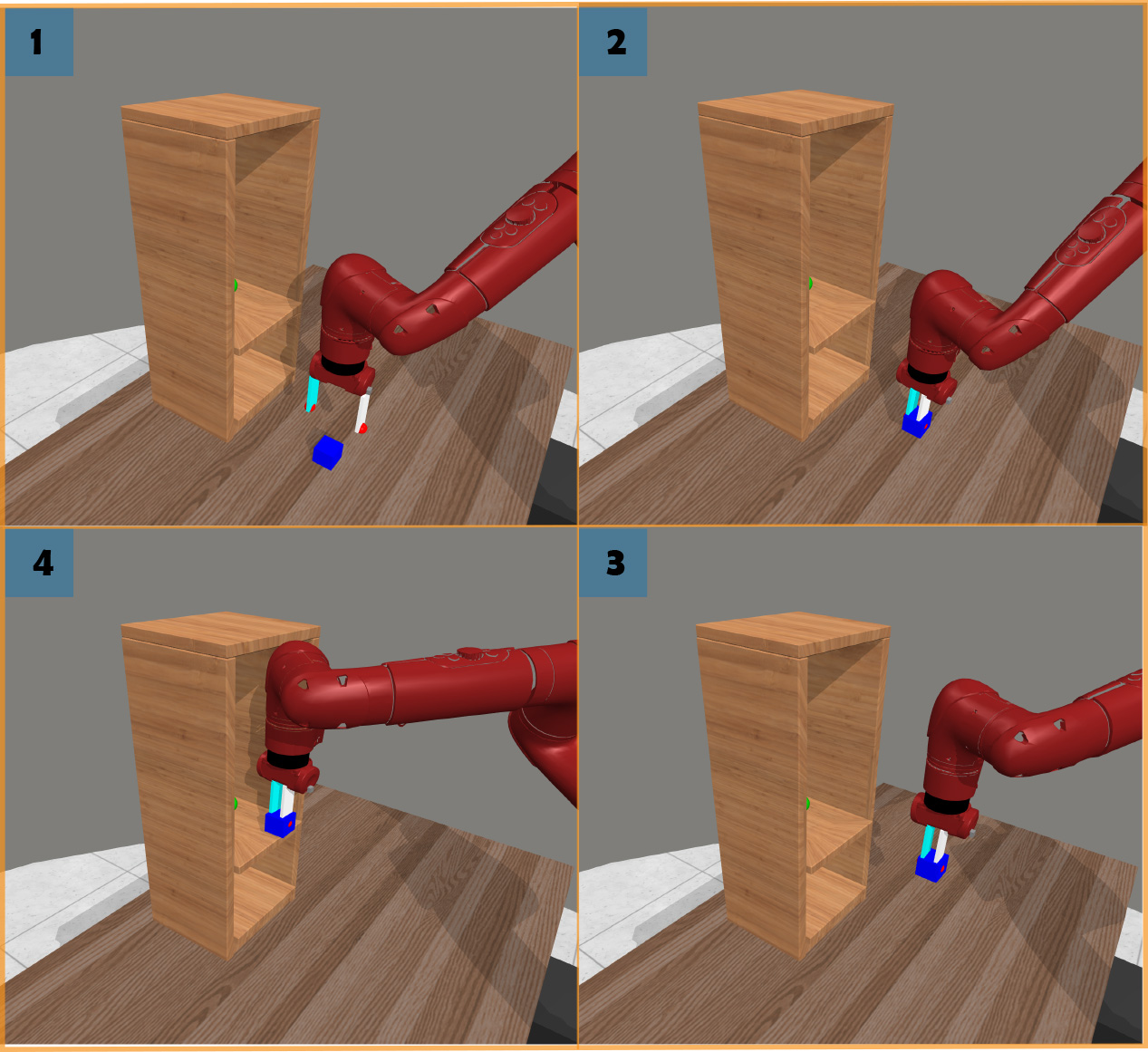}
  \caption*{\textbf{\textit{Metaworld-Shelf Place}}}
  \label{fig:shelfplace}
\end{minipage}
\caption{\textbf{Visualization of Representative Simulation Task Trajectories}.}
\label{fig:simviz}
\end{figure*}
\subsection{Setup}
\textbf{Benchmarks.} \model and baselines are evaluated across three simulation benchmarks, comprising a total of \textbf{11 distinct tasks}:

\begin{itemize}
\item \textit{\textbf{Adroit}}\cite{Adroit}: It employs a multi-fingered Shadow robot within the MuJoCo\cite{mujoco} environment to execute highly dexterous manipulation tasks. These tasks encompass interactions with both articulated objects and rigid bodies.
\item \textit{\textbf{DexArt}}\cite{dexart}: It leverages the Allegro robot within the SAPIEN\cite{sapien} environment to perform high-precision dexterous manipulation, primarily focusing on tasks involving articulated object manipulation.
\item \textit{\textbf{MetaWorld}}~\cite{metaworld}: It operates primarily within the MuJoCo environment, utilizing a gripper to perform manipulation tasks involving both articulated and rigid objects. It covers a diverse range of skills that are categorized by difficulty levels: easy, medium, hard, and very hard. We tested tasks covering all difficulties except easy.
\end{itemize}

\textbf{Baselines.} Diffusion Policy~\cite{DP} and 3D Diffusion Policy~\cite{DP3} are selected as the 2D and 3D baselines, respectively, for simulation environment evaluations. Given the primary contribution of this work lies in the design of the policy's action head (i.e., the autoregressive dense process), the visual backbone is held constant across all compared methods (ResNet18 for 2D as in Diffusion Policy and MLP for 3D as in 3D Diffusion Policy). 
Specifically, the diffusion process-based action head in the baselines is replaced with the proposed dense process, thereby isolating and fairly evaluating the performance gains attributable only to the dense policy.

To maintain experimental parity, \model and the baseline methods are trained using an identical set of expert demonstrations, with an equivalent number of training iterations. Furthermore, during the deployment phase, both methodologies are subjected to the same number of observation and inference steps.

\textbf{Demonstrations.} Expert demonstrations are generated via script policies for MetaWorld, the VRL3~\cite{vrl3} agent for Adroit, and the PPO~\cite{ppo} agent for DexArt. The mean success rates of these demonstrations are 98.7\%, 72.8\%, and approximately 100\%, respectively. Training is performed using 10 demonstrations for both Adroit and MetaWorld, and 100 demonstrations for DexArt.

\textbf{Protocols.} Adhering to the protocol established in \cite{DP3}, each experiment is executed across three independent trials, utilizing seed values of 0, 1, and 2. For each seed, the policy is evaluated over 20 episodes every 200 training epochs, and the mean of the top 5 success rates is computed. The final reported performance consists of the mean and standard deviation of these success rates across the three seeds.
\begin{figure*}[!t] 
  \centering
  \begin{minipage}{0.24\textwidth}
    \centering
    \includegraphics[width=\textwidth]{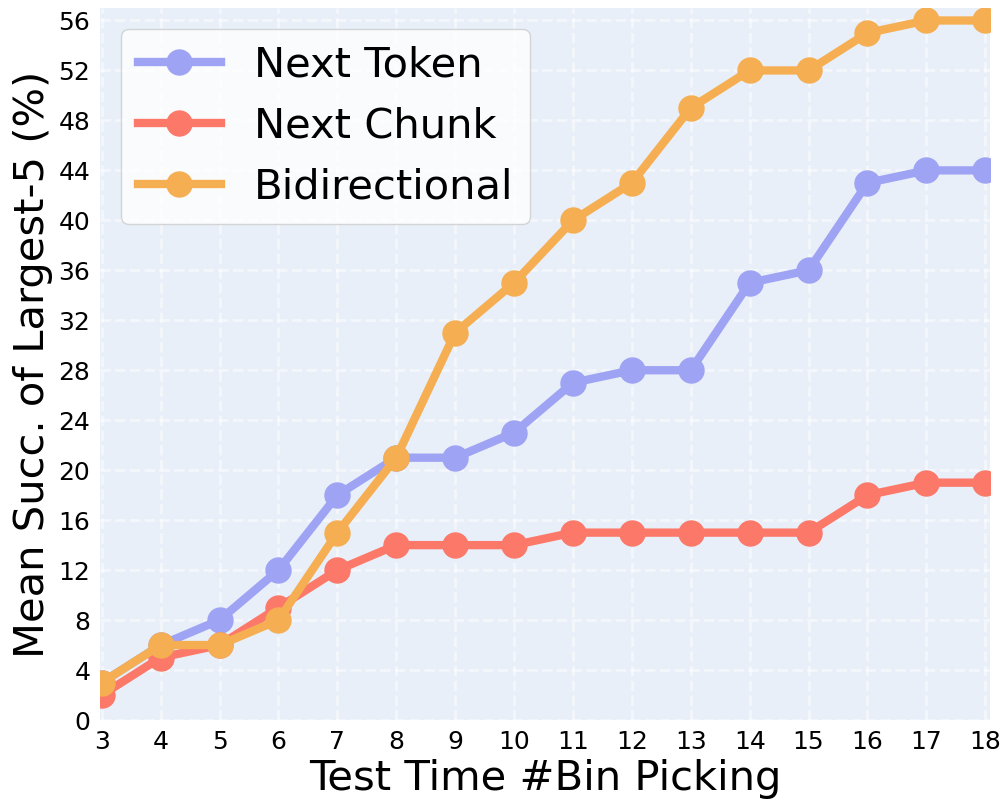} 

  \end{minipage}\hfill
  \begin{minipage}{0.24\textwidth}
    \centering
    \includegraphics[width=\textwidth]{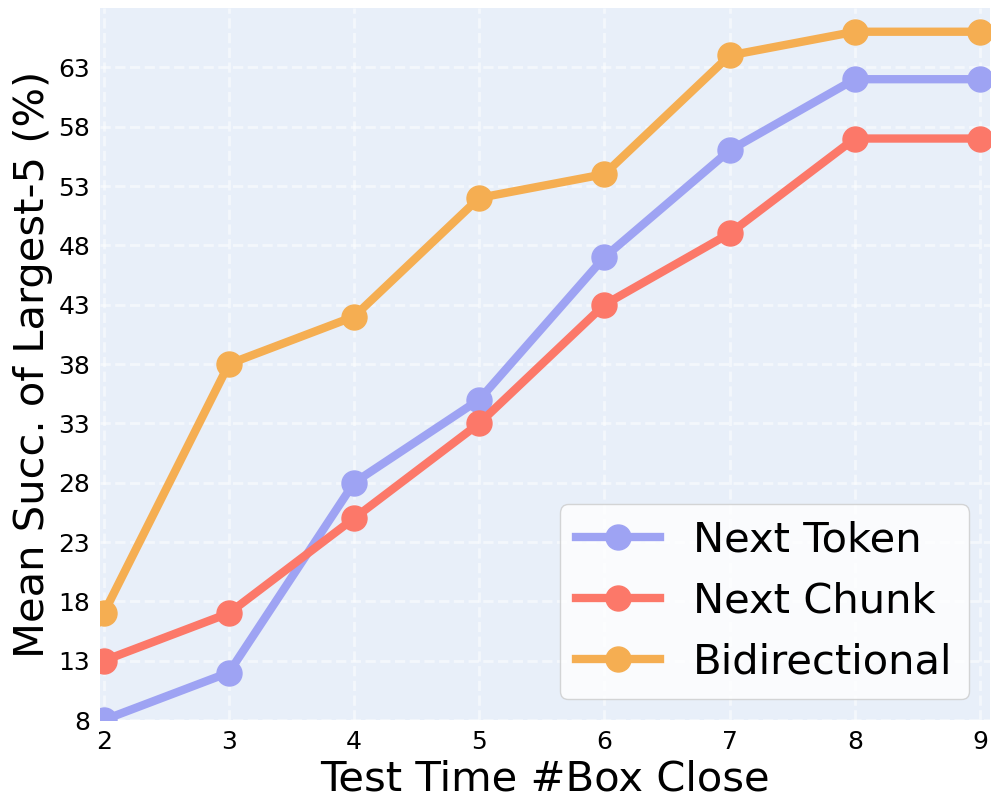}  

  \end{minipage}\hfill
    \begin{minipage}{0.24\textwidth}
    \centering
    \includegraphics[width=\textwidth]{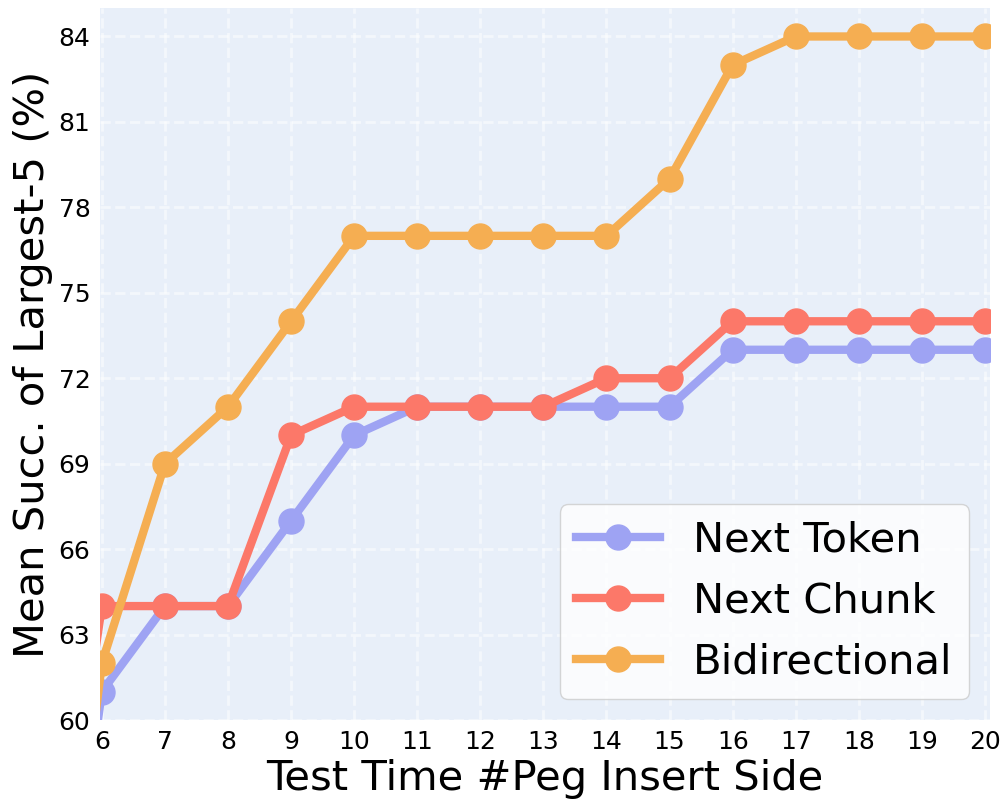}  
  \end{minipage}
  \begin{minipage}{0.24\textwidth}
    \centering
    \includegraphics[width=\textwidth]{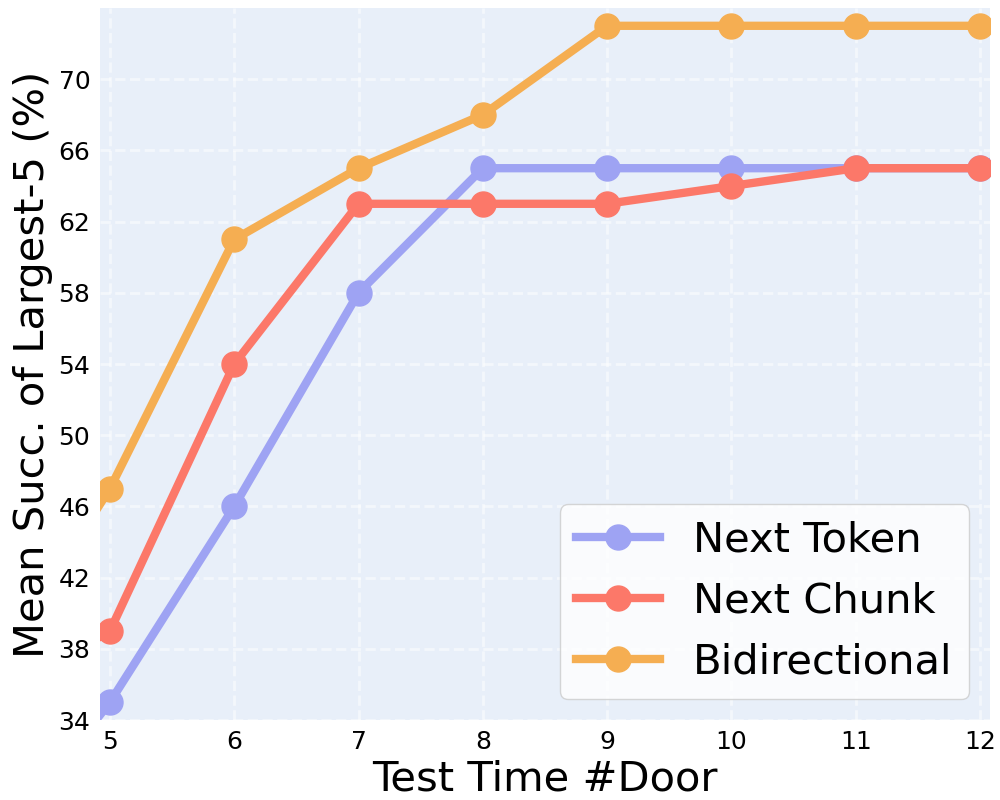}  
  \end{minipage}
  \caption{\textbf{Learning efficiency of different autoregressive paradigm across four different tasks.} The x-axis is the ID of the test time point; The y-axis records the mean of top-5 success rates at the current test time point.}
  \label{simablation}
  \vspace{-10pt}
\end{figure*}
\subsection{Results}
The experimental results, as presented in \cref{simtask}, demonstrate that \model consistently outperforms the baselines across a majority of tasks in both 2D and 3D environments. Specifically, \model exhibits a \textbf{19\% higher} success rate compared to DP3, and a \textbf{27\% higher} success rate compared to DP. 
The task completion trajectories of \model on representative simulation tasks are visualized in \cref{fig:simviz}.

The most significant performance gains are observed in tasks such as shelf place and bin picking, which are representative of object pose estimation; pen, which exemplifies high-DoF dexterous manipulation; and Peg insert side, which is representative of contact-rich manipulation. These results indicate that \model, with an identical visual backbone, exhibits superior downstream capabilities, including more precise estimation of manipulated object states and more robust motion planning. Operations such as \textit{Rotating a pen} and \textit{Peg Insert Side} are particularly sensitive to errors over long horizons, while the bidirectional modeling of action sequences afforded by the \model enables smoother and more coherent actions, thereby substantially increasing the success rate.

\subsection{Ablation}
Ablation studies are conducted on 4 challenging manipulation tasks—\textit{Door, Bin Picking, Shelf Place, Box Close}—which demand precise object localization and high-DoF articulated object manipulation. These experiments specifically targeted different autoregressive paradigms. We altered the \model from a bidirectional architecture to a unidirectional, autoregressive learning framework, exploring both next-token prediction and next-chunk prediction. Given the limited number of steps in the simulation environment, the chunk size was set to 2. Following the established evaluation protocol, we recorded the mean of the top-5 success rates at each test point (until convergence) for each of the three paradigms.

As depicted in Fig~\ref{simablation}, bidirectional prediction exhibited superior learning efficiency on challenging tasks, achieving favorable performance within a shorter time frame and demonstrating a higher ceiling for success rates. This is attributed to the heightened importance of temporal action coherence in tasks with lower error tolerance. The bidirectional token-aware learning approach facilitates the generation of more coherent and fluid action sequences.

\section{Real-World Experiments}
\begin{table*}[!t]
    \centering
    \renewcommand\arraystretch{1.6}
    \setlength\tabcolsep{3pt}
    \setlength{\aboverulesep}{0pt}
    \setlength{\belowrulesep}{0pt}
    \footnotesize
    \begin{tabular}{c|cccccccccccc}
        \hline
        \multirow{2}{*}{\textbf{Method}}  & \multicolumn{1}{c}{\textit{\textbf{Put Bread}}} & \multicolumn{1}{c}{\textit{\textbf{Open Drawer}}} & \multicolumn{3}{c}{\textit{\textbf{Pour Balls}}} & \multicolumn{2}{c}{\textit{\textbf{Flower Arrangement}}}& \\
        \cmidrule(r){2-2}\cmidrule{3-3}\cmidrule(l){4-6}\cmidrule(l){7-9}
        & \textit{Succ. (\%)} & \textit{Succ. (\%)} & \textit{Poured (\%)} & \textit{Balls $\uparrow$} & \textit{Complete(\%)} & \textit{Succ.(\%)} & \textit{Flowers $\uparrow$} &  \\
        \hline
        \cellcolor[HTML]{ecf8f8}3D \model &\cellcolor[HTML]{ecf8f8}$\mathbf{85}$ & \cellcolor[HTML]{ecf8f8}$\mathbf{45}$ & \cellcolor[HTML]{ecf8f8}$85$ & \cellcolor[HTML]{ecf8f8}$\mathbf{7.30}/10$ & \cellcolor[HTML]{ecf8f8}$\mathbf{60}$ &\cellcolor[HTML]{ecf8f8}$\mathbf{70}$&\cellcolor[HTML]{ecf8f8}$\mathbf{1.0}/3.0$ \\
        RISE~\cite{RISE} & $75$ & $40$ & $\mathbf{95}$ & $6.85/10$ & $25$ &$50$&$0.6/3.0$ \\
        \hline
        \cellcolor[HTML]{ecf8f8} 2D \model & \cellcolor[HTML]{ecf8f8}$\mathbf{55}$ & \cellcolor[HTML]{ecf8f8}$\mathbf{20}$ & \cellcolor[HTML]{ecf8f8}$\mathbf{35}$ & \cellcolor[HTML]{ecf8f8}$\mathbf{3.30}/10$ & \cellcolor[HTML]{ecf8f8}$\mathbf{25}$ &\cellcolor[HTML]{ecf8f8}$-$&\cellcolor[HTML]{ecf8f8}$-$\\
        Diffusion Policy~\cite{DP} & $40$ & $\mathbf{20}$ & $30$ & $2.35/10$ & $20$ &$-$&$-$ \\
        ACT~\cite{ACT} & $35$ & $10$ & $30$ & $2.75/10$ & $20$ &$-$&$-$ \\
        \hline
    \end{tabular}
    \caption{\textbf{Detailed performance of \model\ and baselines in real-world tasks.}}
    \label{realtask}
\end{table*}

\begin{figure*}[!t]
\centering
\begin{minipage}{0.245\textwidth}
  \centering 
  \includegraphics[width=\textwidth, height=3.2cm]{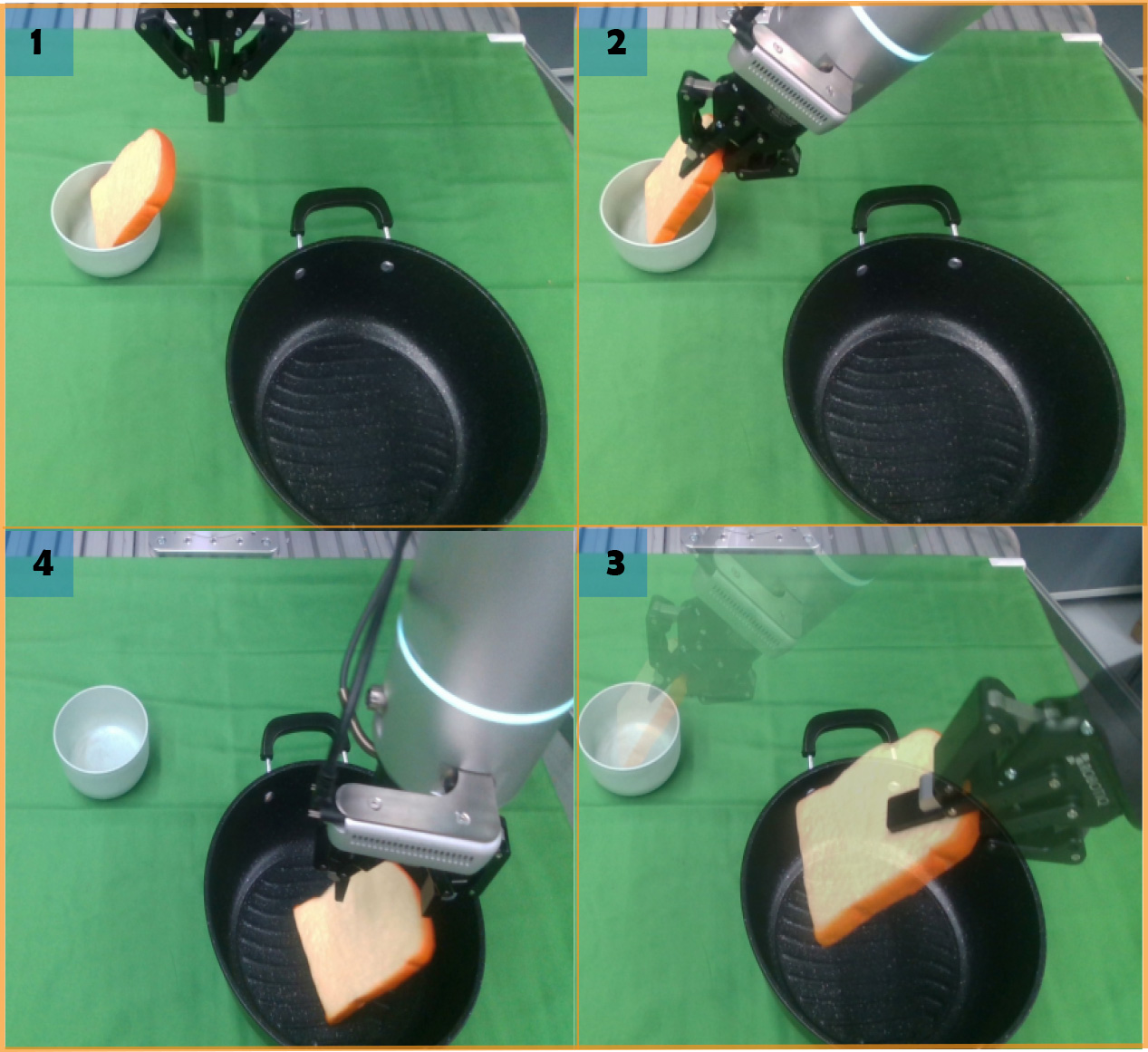}
  \caption*{\textbf{\textit{Put Bread into Pot}}}
  \label{fig:bread}
\end{minipage}\hfill
\begin{minipage}{0.245\textwidth}
  \centering 
  \includegraphics[width=\textwidth, height=3.2cm]{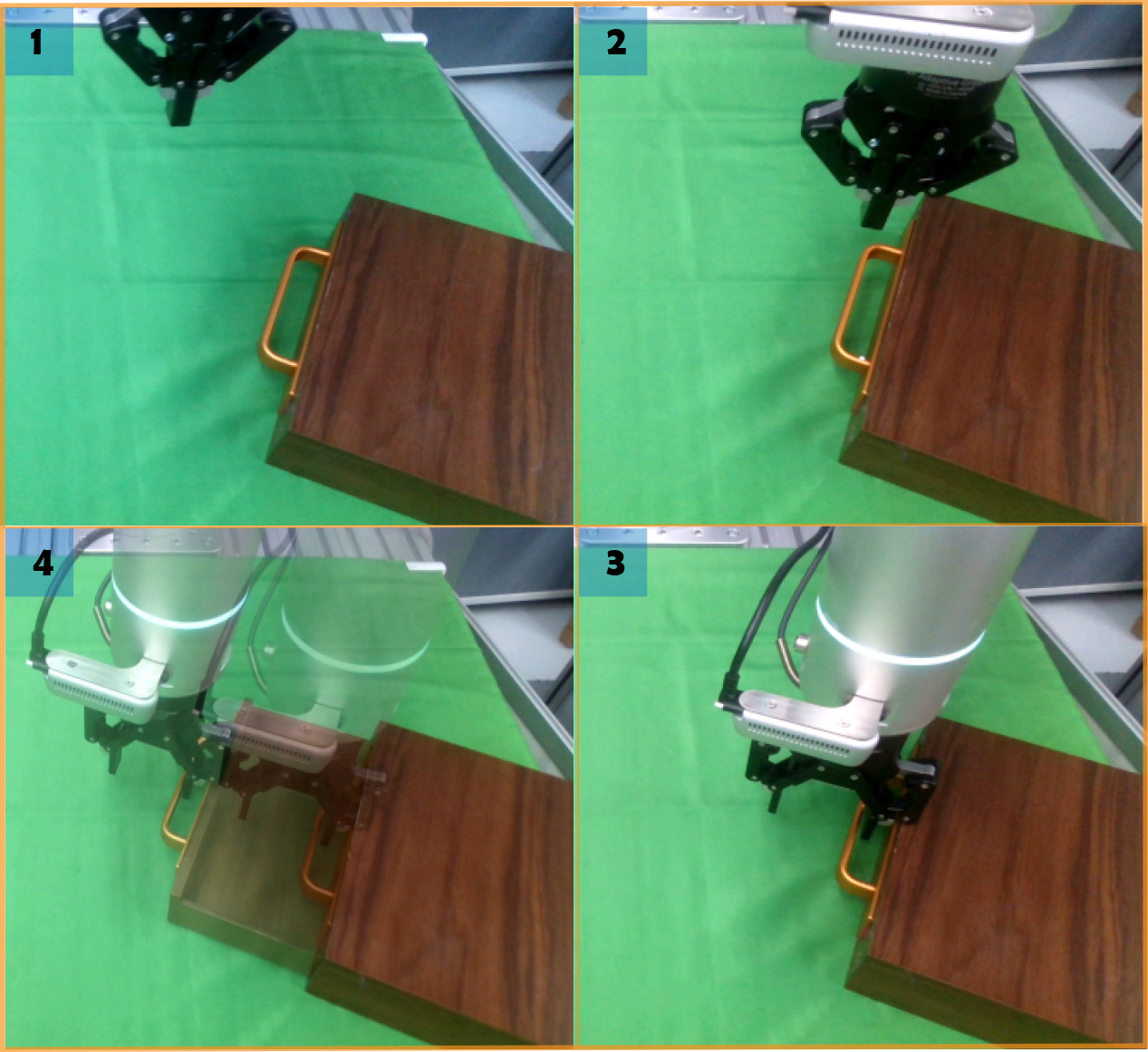}
  \caption*{\textbf{\textit{Open Drawer}}}
  \label{fig:drawer}
\end{minipage}\hfill
\begin{minipage}{0.245\textwidth}
  \centering 
  \includegraphics[width=\textwidth, height=3.2cm]{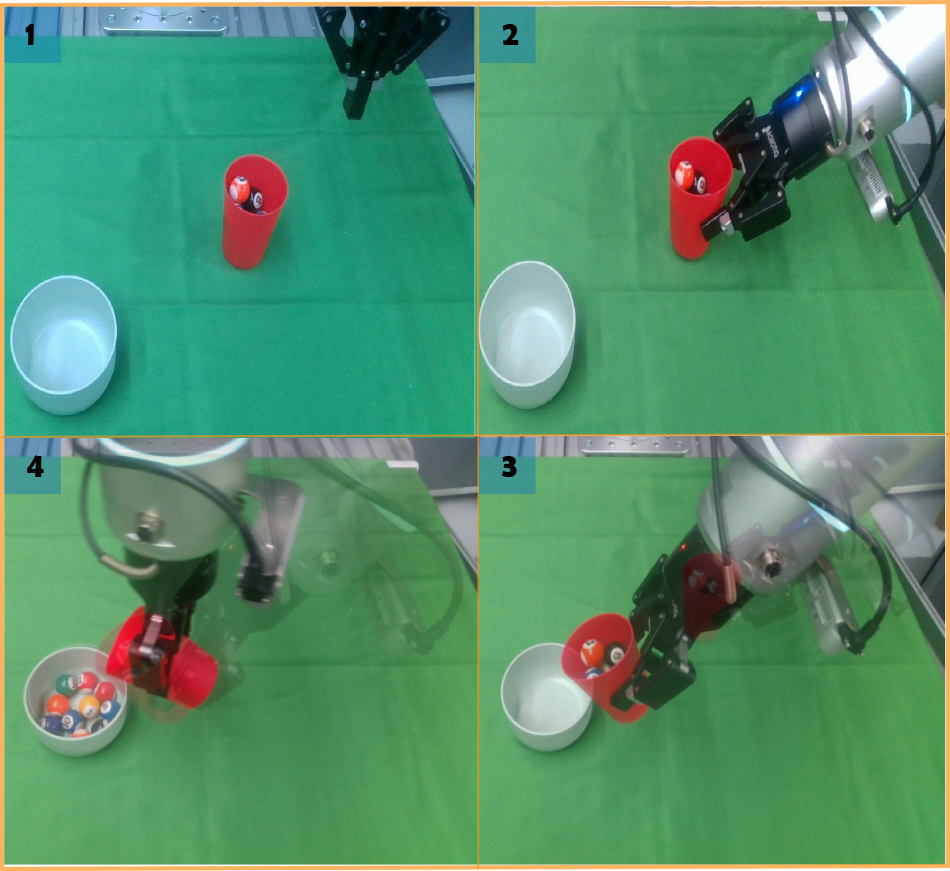}
  \caption*{\textbf{\textit{Pour Balls}}}
  \label{fig:pour}
\end{minipage}\hfill
\begin{minipage}{0.245\textwidth}
  \centering 
  \includegraphics[width=\textwidth, height=3.2cm]{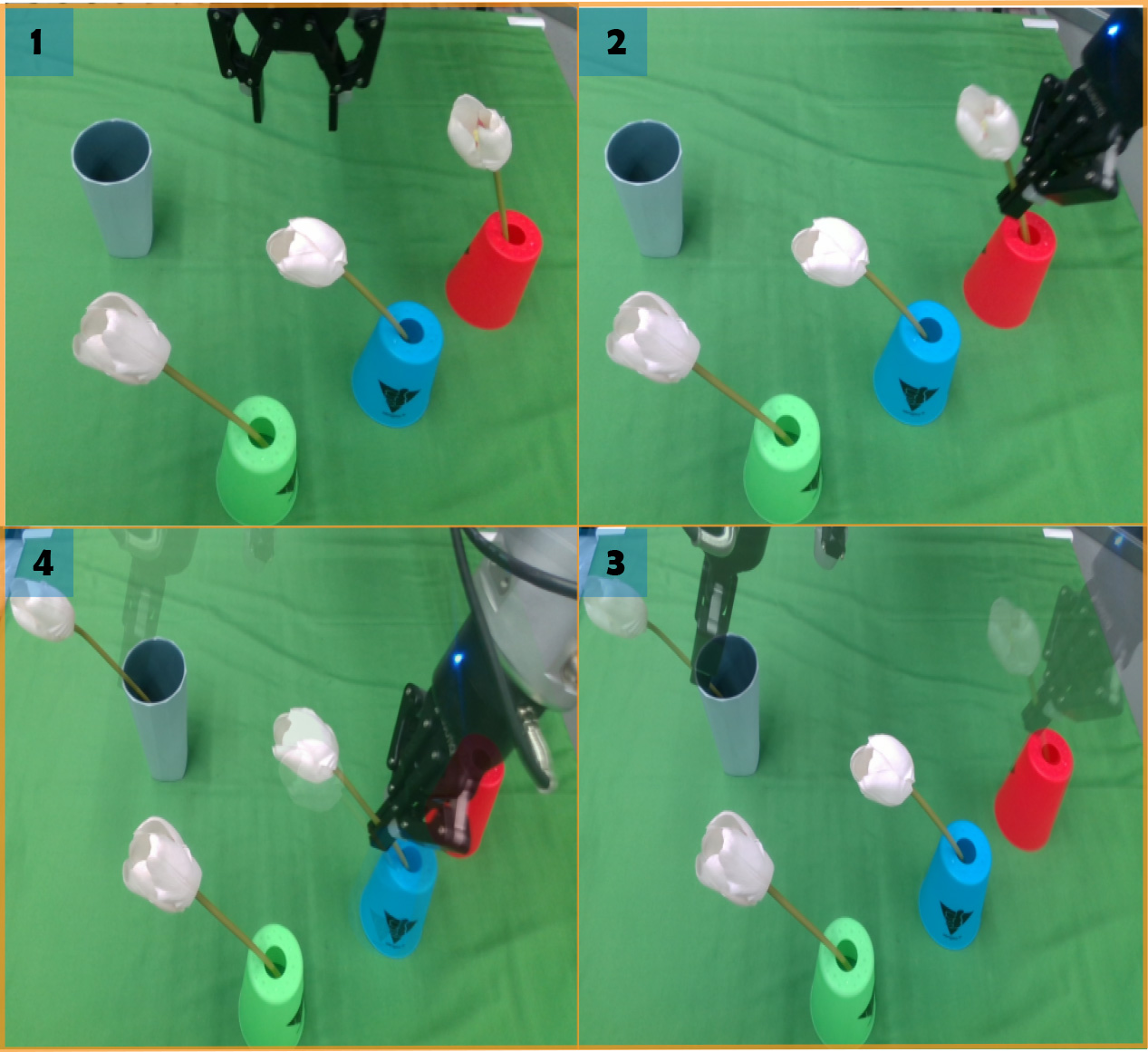}
  \caption*{\textbf{\textit{Flower Arrangement}}}
  \label{fig:flower}
\end{minipage}
\caption{\textbf{Visualization of representative real-world task trajectories.}}
\end{figure*}
\subsection{Setup}

\textbf{Platform.} For real-world experiments, we utilize a Flexiv Rizon robotic arm equipped with a Robotiq 2F-85 gripper for object manipulation. A stationary top-down Intel RealSense D415 RGB-D camera provides 3D perception by capturing single-view global workspace point clouds. As for multi-view 2D perception,  the above top-down camera provides a global RGB image, and an in-hand Intel RealSense D435 RGB-D camera provides a local RGB image. The robot workspace is defined as a 40 cm \(\times\) 60 cm rectangular area in front of the robot. All hardware components are connected to a workstation featuring an Intel i9-10980XE CPU and an NVIDIA 2080 Ti GPU, which facilitates data acquisition and evaluation.

\textbf{Tasks.} We select 4 tasks for experimentation: \textbf{\textit{Put Bread into Pot}} (soft body manipulation), \textbf{\textit{Open Drawer}} (articulated object manipulation),  and \textbf{\textit{Pour Balls}} (6-DoF task) and \textbf{\textit{Flower Arrangement}} (Long-horizon multi-object manipulation).

\textbf{Demonstrations.} For each task, 50 expert demonstrations are acquired via end-effector teleoperation using haptic devices, employing the same setup and procedures as detailed in~\cite{rh20t, RISE, cage, mba}.

\textbf{Baseline.} We select DP~\cite{DP} and ACT~\cite{ACT}, the most widely used methods, as our 2D baselines.
We use RISE~\cite{RISE}, the current state-of-the-art policy for 3D real-world tasks, which is also a diffusion-based policy, as our 3D baseline. 
Consistent with the simulation experiments, the objective is to demonstrate the improvements in manipulation tasks afforded by the Dense Policy's action head.
Therefore, we employ ResNet18 as our 2D visual backbone, as utilized in DP.
Concurrently, we adopt the sparse convolutional networks~\cite{spatioenc} introduced by RISE as the 3D visual backbone, which is more suitable for real-world tasks.

\textbf{Protocols.} For the real-world evaluation, 20 trials are performed per method for each task, unless otherwise specified. All methods are evaluated under closely matched randomized initial scene configurations for each trial. All models, including \model, are trained for 1000 epochs, except for ACT. Due to the need for more extensive training to achieve optimal performance with the CVAE~\cite{cvae}, we follow the official ACT recommendations and train it for 2000 epochs.

\subsection{Performance}
\qheading{\textit{Put Bread into Pot}} is a pick-and-place task involving soft-body objects, presenting a significant challenge due to the policy's robustness requirements against potential bread deformation upon contacts. 

\model outperforms existing baselines in \cref{realtask}, particularly exhibiting a substantial improvement in 2D scenarios. We observed that \model demonstrates a higher probability of adjusting its position during the pick and place processes when localization errors occur. We attribute this to the more comprehensive modeling of sequential dependencies, leading to the generation of less rigid and more robust actions.

\qheading{\textit{Open Drawer}} is a two-stage manipulation that necessitates accurate grasping of the drawer handle followed by a horizontal pulling motion to open the drawer. The primary challenge arises from the minimal clearance between the handle and the drawer surface; even minor pose estimation inaccuracies can lead to grasping failures or gripper slippage during the pulling phase. The task is highly sensitive to the policy's prediction accuracy. 

In this task, \model demonstrates comparable performance to Diffusion-Based Models in 2D and surpasses ACT. In 3D, \model outperforms Diffusion-Based approaches, as detailed in \cref{realtask}. We prove that \model provides more precise action predictions compared to diffusion models, thereby facilitating challenging articulated object manipulation.

\qheading{\textit{Pour Balls}} is a 6-DoF manipulation task. This task entails a robot arm lifting a cup containing 10 balls and subsequently pouring them into a bowl. The inherent difficulties are twofold:
\begin{enumerate}[label=\textbf{\roman*)},left=-10pt]
    \item The cup's varying diameter along its height necessitates adaptive gripper control; the robot must maintain a secure grasp throughout the pouring motion to prevent slippage and ensure proper rotation. This requires a generalizable strategy for determining the appropriate gripper aperture at each height. 
    \item The 6-DoF nature of the task demands simultaneous and precise control of both rotational and translational movements. This comprehensively evaluates the policy's ability to learn across all action dimensions, mitigating potential biases towards specific degrees of freedom.
\end{enumerate}

\begin{figure*}[!t] 
  \centering
  \begin{minipage}{0.35\textwidth}
    \centering
    \includegraphics[width=\textwidth]{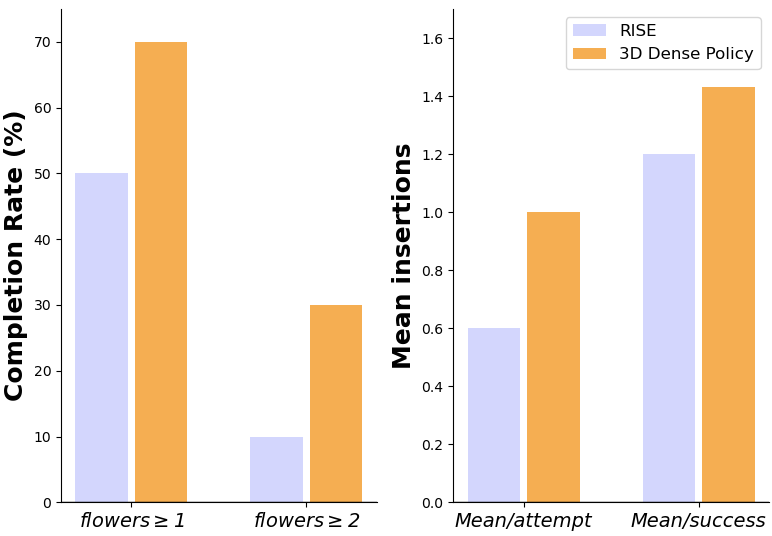}  
    \caption{Comparison of RISE and \model in the \textit{\textbf{Flower Arrangement}} task.}
    \label{flower_comp}
  \end{minipage}\hfill
  \begin{minipage}{0.31\textwidth}
    \centering
    \includegraphics[width=\textwidth]{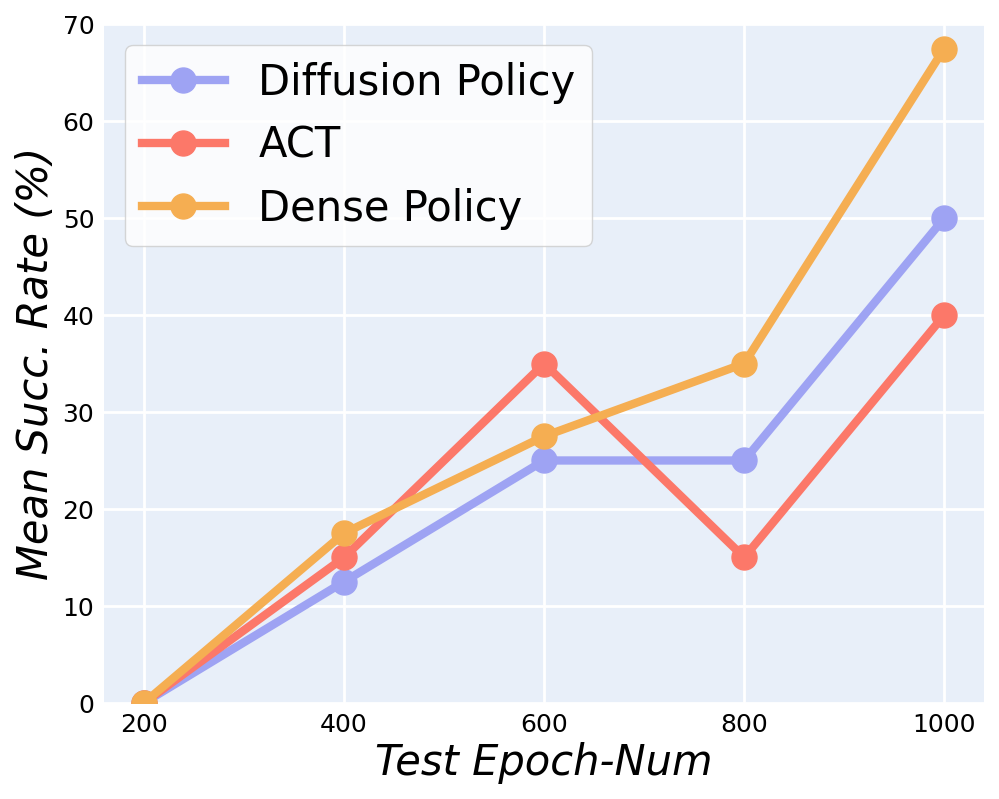} 
    \caption{Learning Efficiency of three policy models in real-world experiments.}
    \label{succcomp}
  \end{minipage}\hfill
  \begin{minipage}{0.31\textwidth}
    \centering
    \includegraphics[width=\textwidth]{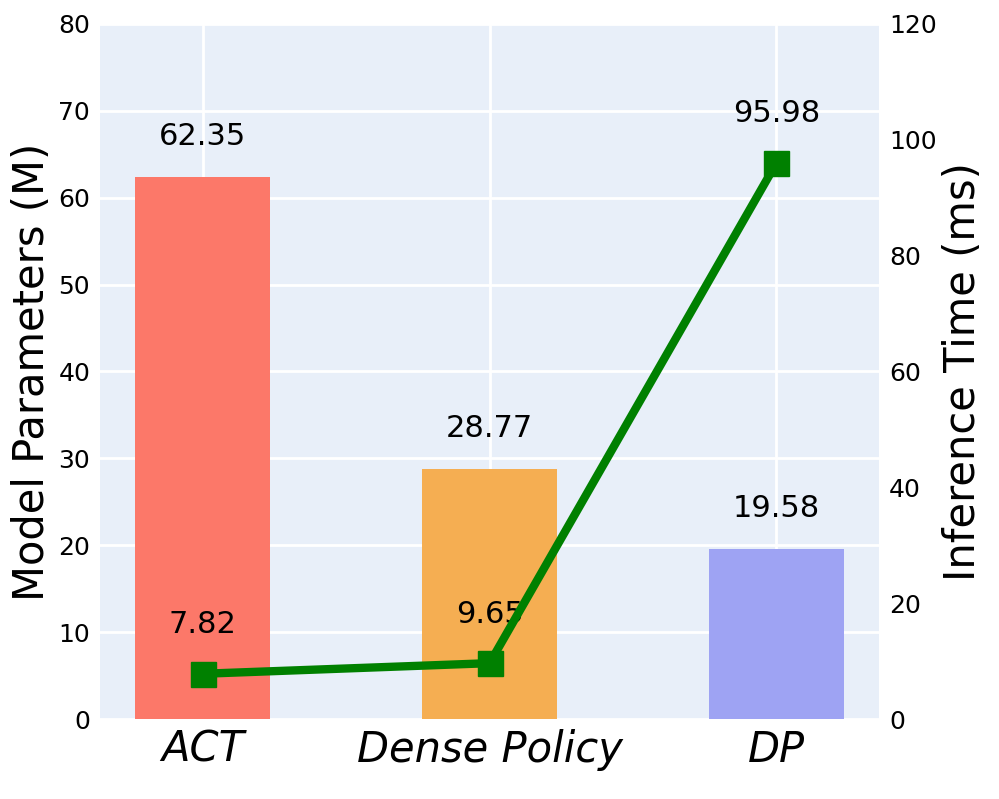}  
    \caption{Inference time \& number of parameters of action heads.}
    \label{infcomp}
  \end{minipage}\hfill
\end{figure*}

As detailed in \cref{realtask}, the performance evaluation is conducted based on the following metrics: \textit{Poured}: the success rate of pouring at least one ball into the bowl; \textit{Balls}: the average number of balls successfully poured into the bowl per trial; \textit{Complete}: the success rate of pouring all 10 balls into the bowl. 

\model achieves the best performance in key pouring metrics, specifically the average number of balls poured and the complete transfer of all balls. We posit that this superior performance stems from \model's learned ability to execute smooth, fluid pouring motions, minimizing deflection errors. This contrasts favorably with diffusion-based policies, which are more susceptible to pouring deviations that can hinder complete ball transfer. However, in 3D benchmarks,  \model shows a slightly reduced success rate of pouring at least one ball compared to RISE, primarily due to occasional instances of constantly excessive gripper tightness, which indicates a potential area for improvement in adaptive error correction.

\qheading{\textit{Flower Arrangement}} is a long-horizon, multi-object manipulation task that requires sequentially inserting three flowers, initially located at random positions, into a vase. The key challenges lie in the extended task horizon and the necessity, in certain configurations, to retrieve flowers in a specific order to avoid collisions with other flowers. This necessitates robust spatial reasoning capabilities within the model to successfully navigate the complex task procedure.

Our experiments reveal that existing 2D policies, including \model, lack the requisite complex spatial reasoning abilities, primarily due to the limitations of 2D representations in capturing intricate spatial relationships. In 3D scenarios, as shown in \cref{realtask}, \model significantly outperforms RISE. 

Specifically, \model achieves a 20\% higher success rate in completing at least one flower insertion and increases the average number of flowers inserted per trial by 0.4. Furthermore, as shown in \cref{flower_comp}, when considering trials where more than one flower is successfully inserted, RISE achieves a 10\% success rate, whereas \model demonstrates a substantial success rate of 30\%. Considering all test cases (reported as \textit{Mean/attempt} in \cref{flower_comp}), \model achieves an average of 1.0 flower insertions per trial, while RISE averages 0.6.
When analyzing only successful insertion trials (excluding complete failures, reported as \textit{Mean/success} in \cref{flower_comp}), \model averages 1.43 flower insertions compared to RISE's average of 1.20,
highlighting its superior long-horizon task execution capabilities.

\subsection{Efficient \model}
\qheading{Ease of Training.} We take the view that \model offers improved trainability compared to existing generative policies. Diffusion models typically necessitate a high number of iterative steps to ensure accurate modeling of the sample distribution. Furthermore, VAE-based policies, in particular, require extensive optimization due to the variational process involving encoding and decoding. The latent variables (or codebook in the case of VQ-VAE) introduce an additional optimization objective, demonstrably increasing the training burden. We evaluate the success rate (ratio of total balls poured to the theoretical maximum) of \textit{pour balls} tasks in 2D scenarios with simple initial deployment, using models trained for every 200 epochs up to 1,000 epochs. The results, presented in \cref{succcomp}, demonstrate that ACT's training is unstable and its final performance is inferior to both DP and \model. Furthermore, DP exhibits lower training efficiency and final performance compared to \model, corroborating our hypothesis.

\qheading{Light weight and Rapid inference.} We quantify the parameter count of the action head (excluding the vision backbone) and the inference time of each action head (processing information already encoded by the vision backbone) for each policy. The results are presented in \cref{infcomp}. Notably, \model achieves comparable inference speed to ACT with less than half the number of parameters. While \model has 9.19M more parameters than DP, it achieves an inference speed nearly ten times faster. Our iterative prediction maintains a high inference speed, owing to the logarithmic recursive process, as demonstrated in~\cref{relat}. Thus we highlight that \model effectively balances lightweight design and rapid inference, and furthermore, it outperforms both ACT and DP in terms of performance.

\section{Conclusion}
This paper introduces Dense Policy, an autoregressive policy that achieves efficient imitation learning and high-quality coarse-to-fine action generation through bidirectional sequence expansion.
The proposed action head demonstrates robust task performance across various visual input modalities and is compatible with diverse visual backbones.
Furthermore, it is demonstrated to be lightweight and computationally efficient during inference, alongside enhanced efficiency and stability during training.
A limitation of this work lies in the unexplored potential of extending Dense Policy into a more general-purpose autoregressive VLAs, and its stability when scaled up to larger foundation models to challenge Diffusion Head-based VLAs.
Future work will focus on investigating and developing the potential of Dense Policy in addressing these challenges.
\clearpage

{
    \small
    \bibliographystyle{ieeenat_fullname}
    \bibliography{main}
}

\end{document}